\documentclass[conference]{IEEEtran}
\IEEEoverridecommandlockouts
\usepackage{cite}
\usepackage{amsmath,amssymb,amsfonts}
\usepackage{algorithmic}
\usepackage{graphicx}
\usepackage{textcomp}
\usepackage{xcolor}
\def\BibTeX{{\rm B\kern-.05em{\sc i\kern-.025em b}\kern-.08em
    T\kern-.1667em\lower.7ex\hbox{E}\kern-.125emX}}
\begin{document}

\title{AI-based approach for improving the detection of blood doping in sports\\
}

\author{\IEEEauthorblockN{Maxx Richard Rahman}
\IEEEauthorblockA{\textit{Saarland University} \\ 
Germany}
\and
\IEEEauthorblockN{Jacob Bejder}
\IEEEauthorblockA{\textit{University of Copenhagen} \\
Denmark}
\and
\IEEEauthorblockN{Thomas Christian Bonne}
\IEEEauthorblockA{\textit{University of Copenhagen} \\
Denmark}
\and
\IEEEauthorblockN{Andreas Breenfeldt Andersen}
\IEEEauthorblockA{\textit{University of Copenhagen} \\
Denmark}
\and
\IEEEauthorblockN{Jesús Rodríguez Huertas}
\IEEEauthorblockA{\textit{University of Granada} \\
Spain}
\and
\IEEEauthorblockN{Reid Aikin}
\IEEEauthorblockA{\textit{World Anti-Doping Agency} \\
Canada}
\and
\IEEEauthorblockN{Nikolai Baastrup Nordsborg}
\IEEEauthorblockA{\textit{University of Copenhagen} \\
Denmark}
\and
\IEEEauthorblockN{Wolfgang Maaß}
\IEEEauthorblockA{\textit{Saarland University} \\
Germany}
}

\maketitle

\begin{abstract}
Sports officials around the world are facing incredible challenges due to the unfair means of practices performed by the athletes to improve their performance in the game. It includes the intake of hormonal based drugs or transfusion of blood to increase their strength and the result of their training. However, the current direct test of detection of these cases includes the laboratory-based method, which is limited because of the cost factors, availability of medical experts, etc. This leads us to seek for indirect tests. With the growing interest of Artificial Intelligence in healthcare, it is important to propose an algorithm based on blood parameters to improve decision making. In this paper, we proposed a statistical and machine learning-based approach to identify the presence of doping substance rhEPO in blood samples.
\end{abstract}


\section{Introduction}
Artificial Intelligence (AI) has shown potential improvement in the sports industry, whether to identify players' unique talents, detect previous injuries, or even assist decision-making. Automated Sports Journalism is a good example where AI is used in guiding sports journalism through automation. Automated Insights, an AI-driven platform that translates data from sports leagues into stories by using natural language, increases the media's reporting capacity \cite{bib1}. Not only the application is boundless to the development of sports, but it can also be used to ensure fairness in sports by athletes.

Athletes have a desire to increase their physical performance to obtain better results which leads some of them to seek alternative methods. Therefore, blood doping practices in sports have been around for several decades. The main reason for the various forms of blood doping is that they are easy to perform, and the detection is difficult. Blood doping can be performed by various methods like through blood transfusion along with the intake of erythropoietic stimulants or hormones such as recombinant human erythropoietin (rhEPO), the transfusion of haemoglobin-based substitutes to increase the number of red blood cells (RBC) or enhance oxygen transfer to the muscles through a modification of athlete's oxygen transport capacity \cite{bib2}. World Anti-Doping Agency (WADA), along with the law enforcement authorities, have initiated the 'World Anti-Doping Program' to describe all the prohibited substances that can be used as doping drugs among athletes and their effect on the performance of the athletes \cite{bib3}.

One of the substances is rhEPO, a recombinant-based treatment that increases erythropoiesis which has the ability to increase oxygenation in the blood. Due to the fact that rhEPO is not easily differentiable from the naturally occurring erythropoietin, it is commonly misused by athletes in their training \cite{bib4}. WADA is responsible for identifying the athletes who take rhEPO treatment. The fundamental distinction can be made between direct and indirect methods of the detection of rhEPO in blood. The direct methods meet the medico-legal requirements of WADA in which the presence of rhEPO is inspected in the blood or urine samples using laboratory-based methods. Some methods are RNA testing, the biomarker test, immunoassays, liquid chromatography, mass spectrometry, etc. However, all these methods require experts to have the proper domain knowledge to collect and analyse the blood samples according to the regulations. Moreover, expert analysis of blood samples includes time and cost factors that also need to be considered in the overall decision-making process. These limitations lead us to investigate the indirect methods of detection. Since rhEPO intake produces characteristic changes in haematological parameters, it is possible to authenticate athletes based on indirect indicators for blood doping.

In this paper, we started by reviewing the literature on indirect detection methods with an emphasis on statistical methods. Then, we presented the procedure of the clinical experiment we conducted to collect the data. Next, this data is analyzed by using statistical methods. Then, we detail the three machine learning algorithms that we trained to identify doping cases and also describe the evaluation of our algorithms. Finally, we show the performance of the algorithms and discuss the results with possible future research.

\section{Related Work}\label{sec1}

We briefly review the related works in terms of indirect methods for the detection of rhEPO in blood. Studying the effect of rhEPO in blood by statistical approaches is not new in the doping community. \cite{bib5} proposed the statistical-based software called A.R.I.E.T.T.A., which can analyse the haematological profiles of athletes to detect abnormal patterns. A risk score is calculated based on different parameters by considering the shift on the present value from the reference values based on the number of standard deviations. Another software GASepo proposed by \cite{bib5a} aims at the quantitative analysis of images obtained by electrofocusing and double-blotting techniques. It is based on the image segmentation of individual bands that are used to detect rhEPO in doping controls. 

\cite{bib6} compares different machine learning algorithms like Support Vector Classification, Naive Bayes, Logistic Regression with different resampling techniques like TOMEK and SMOTE methods to identify the doping activities of the athletes. They analysed the blood samples of 791 UFC fighters. The results obtained suggest that support vector classification and logistic regression combined with oversampling could be an effective method for identifying doping cases. Furthermore, they achieved a sensitivity of 44\% with an accuracy of 73\% in their results.

\cite{bib7} introduced the Abnormal Blood Profile Score (ABPS), which is an indirect and universal test based on the statistical classification of indirect biomarkers of altered erythropoiesis. The calculation of ABPS is based on Support Vector Machine and Naive Bayes algorithms together with cross-validation techniques to map labelled reference profiles to target outputs. As a result, they achieved a sensitivity of 45\% at 100\% specificity, which is used to comply with the medical and legal standards required by WADA. Therefore, this is the current state-of-the-art (SOTA) method which is used as a baseline for this paper. As an extension, \cite{bib8} in their work developed R-based package $ABPS$ to calculate the Abnormal Blood Profile Score.

The existing work on indirect methods is limited to the statistical analysis and classical machine learning algorithms with one classifier. In this paper, we proposed an ensemble and boosted algorithms where more than one classifier are trained in parallel, and the final prediction is made by the collective decision of the classifiers.

\section{Theory}\label{sec3}

\subsection{Erythropoietin}\label{subsec3.1}
Erythropoietin (EPO) is a peptide hormone naturally secreted by the kidney to stimulate the production of red blood cells in the blood. It increases the blood capacity to transport oxygen which results in increasing of body endurance \cite{bib9}. One way to naturally increase the production of EPO is through altitude training. The body compensates for the reduced oxygen concentration at a high altitude by releasing EPO. Several synthetically produced substances can stimulate endogenous EPO production, like recombinant human erythropoietin (rhEPO). Therefore, rhEPO was added to WADA’s banned doping list in early 1990 \cite{bib10}.

\subsection{Haematological profile}\label{subsec3.1}
The production of red blood cells by using pharmacological means may cause iron-deficient erythropoiesis, which is indicated by the production of low concentration red blood cells \cite{bib9}. This can be observed by a significant change in different blood parameters like haemoglobin concentrations, red blood cell count, haematocrit, the percentage of reticulocytes, etc. Therefore, a haematological profile is formed, consisting of a set of red blood cell parameters called haematological parameters that show significant changes in the use of rhEPO. The haematological profile consists of haemoglobin concentration (HB), haematocrit (HCT), reticulocytes percentage (RET\%), reticulocytes count (RET\#), reticulocytes haemoglobin (RET-HB), mean corpuscular volume (MCV), mean corpuscular haemoglobin mass (MCH), mean corpuscular haemoglobin concentration (MCHC), red blood cell count (RBC), red blood cell distribution width - standard deviation (RDW-SD), red blood cell distribution width - coefficient of variation (RDW-CV), white blood cell count (WBC), immature reticulocyte fraction (IRF), low fluorescence reticulocyte fraction (LFR), medium fluorescence reticulocyte fraction (MFR), high fluorescence reticulocyte fraction (HFR) and Off-HR score (OFF-HR) \cite{bib10,bib11}.
\[ OFF-HR = HB (g/L) - 60*\sqrt{RET\%} \]

\section{Experiment}\label{sec3}

\subsection{Data collection}\label{subsec3.1}

We performed a clinical experiment for 24 weeks consisting of 12 weeks at sea level and 12 weeks at higher altitude. 39 participants of different age groups, sex, size, etc., participated in the experiment at high altitude and 35 participants out of them participated at the sea-level. Each participant performed a regular training workout consisting of a 5000 m trial run and a treadmill test. The time span of 12 weeks was divided into three periods: Baseline (week 1-4), Intervention (week 5-8) and Follow-up (week 8-12). None of the participants was given any doping substituent in the Baseline and Follow-up periods, whereas in the Intervention period, 11 injections were given to all the participants after every other day. 25 participants were given rhEPO injections, whereas 9 participants were given placebo injections in the experiment at sea-level. In the studies at high altitude, 12 were injected with rhEPO and 27 were given placebo injections. Every participant was monitored regularly, and the blood sample of each participant was collected every week. So, the data statistics used in this analysis is summarised in Table \ref{datatab}. The haematological profile of these blood samples is used to study the effect of rhEPO and labelled accordingly.

\renewcommand*\arraystretch{1.5}
\begin{table}[htbp]
\caption{Number of blood samples collected at sea-level and high altitude}
\begin{center}
\begin{tabular*}{21pc}{@{\extracolsep{\fill}} c c c }
\hline
\textbf{Blood samples} & \textbf{Sea-level} & \textbf{High altitude} \\
\hline
Controlled samples (Placebo) & 306 & 410 \\
rhEPO samples & 100 & 48 \\
\hline
Total samples & 406 & 458 \\
\hline
\end{tabular*}
\label{datatab}
\end{center}
\end{table}

\section{Statistical Analysis}\label{subsec3.2}

\subsection{Data preprocessing}\label{subsec3.1}

The haematological profile of each collected blood sample is measured, which consists of 17 haematological parameters. However, some parameters contain missing values which is due to the measurement error. We used a median imputation strategy to account for these cases where the median of all the samples from the same participant at a given altitude is taken into account.

\subsection{Finding best indicators}\label{subsec3.1}

Some parameters are susceptible to the intake of rhEPO and show a significant change in their values, whereas the difference is negligible in other parameters. We used 2 sample Kolmogorov-Smirnov test (K-S test) to identify the best haematological parameters, which show a significant change in their values on the intake of rhEPO. 
The K-S test \cite{bib12} is a standard statistical test for deciding whether a dataset is consistent with another dataset. The maximum difference between the cumulative distribution function of the two datasets (controlled $F_{a}(x)$ and rhEPO $F_{b}(x)$) is calculated by using
\[ D_{a,b} = sup |F_{a}(x) - F_{b}(x)| \]

The null distribution of this statistic is calculated under the null hypothesis that the two datasets are drawn from the same parent distribution. The null hypothesis is rejected at level $\alpha$ if:
\[ D_{a,b} > \sqrt{-\ln{\frac{\alpha}{2}}.\frac{1+\frac{b}{a}}{2b}} \]

We chose $\alpha$ = 0.001 to determine the best indicators at  confidence level of 99.99\%.

\begin{figure}[hbt!]
\centering{\includegraphics[width=7cm]{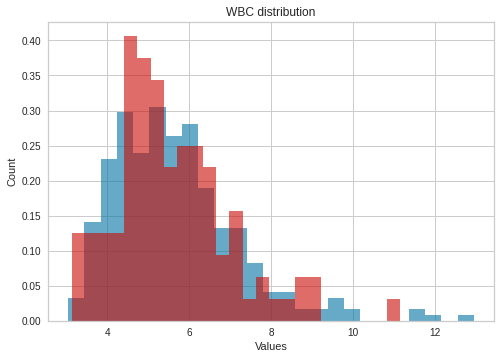}}
\centering{\includegraphics[width=7cm]{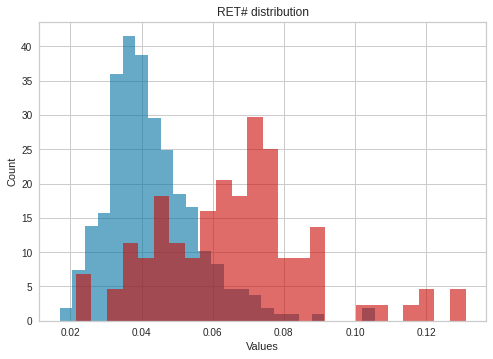}}
\caption{Distribution of WBC and RET\# parameters for controlled (blue) and rhEPO (red) samples at sea-level.\label{kstest}}
\end{figure}

Fig. \ref{kstest} shows the distribution of WBC and RET\# for controlled (blue) and rhEPO (red) samples. WBC has a $p$-value of 0.96, which shows that there is no significant change observed due to the effect of rhEPO. On the other hand, the $p$-value of RET\# 5.6e-16 indicates that it is the best indicator to observe the effect of rhEPO. Similarly, the best 8 parameters are selected based on the $p$-values for both sea-level and high altitude. The $p$-values for all the haematological parameters at sea-level and high altitude are presented in Table \ref{table1} and Table \ref{table2}, respectively.

\begin{table*}[htbp]
\caption{Statistics of haematological parameters for rhEPO and controlled samples at the sea-level (low altitude) including inter-quartile range (IQ1-IQ3), median, mean, min, max and the $p$-value from the K-S test}
\begin{tabular*}{21pc}{@{\extracolsep{\fill}} c c c c c c c c c c c c c c }
\cline{1-14} 
&\multicolumn{2}{c}{\textbf{rhEPO (n=100)}} & & & & & \multicolumn{2}{c}{\textbf{controlled samples (n=609)}} & & & & \multicolumn{2}{c}{\textbf{$p$-value}} \\\cline{2-14} 
Parameter &\multicolumn{1}{c}{mean$\pm$std.} &\multicolumn{1}{c}{min} &\multicolumn{1}{c}{IQ1} &\multicolumn{1}{c}{median} &\multicolumn{1}{c}{IQ3} &\multicolumn{1}{c}{max} &\multicolumn{1}{c}{mean$\pm$std.} &\multicolumn{1}{c}{min} &\multicolumn{1}{c}{IQ1} &\multicolumn{1}{c}{median} &\multicolumn{1}{c}{IQ3} &\multicolumn{1}{c}{max} \\
\cline{1-14}
HB &14.5$\pm$1.2 &11.4 &13.6 &14.4 &15.4 &17.0 &14.2$\pm$1.1 &11.5 &13.4 &14.3 &15.0 &17.0 &4.1e-02 \\
HCT &42.3$\pm$3.3 &32.8 &40.2 &42.1 &44.5 &51.2 &41.2$\pm$3.0 &33.2 &39.3 &41.2 &43.2 &50.1 &7.1e-03 \\
RET\# &0.1$\pm$0.0 &0.0 &0.1 &0.1 &0.1 &0.1 &0.0$\pm$0.0 &0.0 &0.0 &0.0 &0.1 &0.1 &1.3e-15 \\
RET\% &1.4$\pm$0.4 &0.5 &1.1 &1.4 &1.7 &2.7 &0.9$\pm$0.3 &0.3 &0.7 &0.9 &1.1 &2.2 &1.3e-15 \\
RET-HB &33.5$\pm$1.7 &29.4 &32.5 &33.6 &34.5 &36.9 &33.5$\pm$1.8 &27.4 &32.5 &33.5 &34.7 &38.0 &9.7e-01 \\
MCV &89.5$\pm$2.9 &84.2 &87.5 &88.9 &91.8 &96.0 &88.9$\pm$3.2 &81.0 &86.7 &88.6 &91.0 &98.1 &9.2e-02 \\
MCH &30.6$\pm$1.1 &28.3 &29.6 &30.6 &31.2 &32.8 &30.7$\pm$1.4 &26.7 &29.7 &30.7 &31.7 &35.1 &7.1e-01 \\
MCHC &34.2$\pm$0.8 &32.2 &33.6 &34.1 &34.6 &36.5 &34.5$\pm$1.1 &32.0 &33.7 &34.4 &35.0 &38.9 &3.9e-03 \\
RBC &4.7$\pm$0.4 &3.5 &4.5 &4.7 &4.9 &5.7 &4.6$\pm$0.4 &3.6 &4.4 &4.7 &4.9 &5.7 &9.7e-02 \\
RDW-SD &42.5$\pm$2.6 &37.7 &40.4 &42.4 &44.4 &48.4 &41.7$\pm$2.6 &35.2 &40.0 &41.5 &43.1 &54.1 &9.2e-04 \\
RDW-CV &12.9$\pm$0.6 &11.6 &12.4 &12.9 &13.2 &14.4 &12.7$\pm$0.8 &11.6 &12.1 &12.6 &13.0 &17.0 &5.6e-04 \\
WBC &5.6$\pm$1.4 &3.1 &4.6 &5.4 &6.3 &11.1 &5.8$\pm$1.6 &3.0 &4.7 &5.6 &6.7 &14.8 &5.6e-01 \\
IRF &9.9$\pm$3.8 &1.1 &7.4 &9.9 &12.3 &22.7 &6.2$\pm$2.7 &0.0 &4.3 &6.0 &7.9 &15.0 &1.3e-15 \\
LFR &90.1$\pm$3.8 &77.3 &87.7 &90.1 &92.6 &98.9 &93.7$\pm$2.7 &85.0 &92.1 &94.0 &95.6 &99.3 &1.3e-15 \\
MFR &8.6$\pm$3.0 &1.1 &6.7 &8.7 &10.6 &19.0 &5.7$\pm$2.3 &0.7 &4.0 &5.5 &7.1 &13.4 &1.3e-15 \\
HFR &1.4$\pm$0.9 &0.0 &0.7 &1.3 &1.8 &5.7 &0.6$\pm$0.5 &0.0 &0.2 &0.5 &0.9 &3.0 &2.6e-12 \\
OFF-HR &74.3$\pm$15.7 &36.2 &63.7 &73.0 &85.8 &111.6 &85.0$\pm$14.3 &44.8 &75.5 &85.4 &94.8 &119.1 &2.6e-09 \\
\cline{1-14}
\end{tabular*}
\label{table1}
\end{table*}

\begin{table*}[htbp]
\caption{Statistics of haematological parameters for rhEPO and controlled samples at high altitude including inter-quartile range (IQ1-IQ3), median, mean, min, max and the $p$-value from the K-S test}
\begin{tabular*}{21pc}{@{\extracolsep{\fill}} c c c c c c c c c c c c c c }
\cline{1-14} 
&\multicolumn{2}{c}{\textbf{rhEPO (n=48)}} & & & & & \multicolumn{2}{c}{\textbf{controlled samples (n=107)}} & & & & \multicolumn{2}{c}{\textbf{$p$-value}} \\\cline{2-14} 
Parameter &\multicolumn{1}{c}{mean$\pm$std.} &\multicolumn{1}{c}{min} &\multicolumn{1}{c}{IQ1} &\multicolumn{1}{c}{median} &\multicolumn{1}{c}{IQ3} &\multicolumn{1}{c}{max} &\multicolumn{1}{c}{mean$\pm$std.} &\multicolumn{1}{c}{min} &\multicolumn{1}{c}{IQ1} &\multicolumn{1}{c}{median} &\multicolumn{1}{c}{IQ3} &\multicolumn{1}{c}{max} \\
\cline{1-14}
HB &15.1$\pm$1.4 &12.2 &13.6 &15.3 &16.0 &17.6 &14.5$\pm$1.1 &11.9 &13.6 &14.6 &15.5 &16.8 &3.9e-02 \\
HCT &43.6$\pm$3.9 &35.5 &40.2 &44.1 &46.5 &51.5 &41.7$\pm$2.8 &34.7 &39.5 &41.8 &43.9 &47.5 &7.3e-04\\
RET\# &0.1$\pm$0.0 &0.0 &0.1 &0.1 &0.1 &0.1 &0.1$\pm$0.0 &0.0 &0.1 &0.1 &0.1 &0.1 &1.9e-06 \\
RET\% &1.7$\pm$0.5 &0.8 &1.4 &1.7 &2.1 &2.8 &1.3$\pm$0.4 &0.5 &1.0 &1.3 &1.6 &2.9 &6.1e-05 \\
RET-HB &34.7$\pm$1.8 &30.3 &33.6 &34.7 &35.7 &37.7 &34.6$\pm$1.8 &28.4 &33.8 &34.8 &35.8 &38.0 &9.1e-01 \\
MCV &89.9$\pm$2.6 &85.4 &88.3 &89.6 &92.0 &95.9 &87.6$\pm$3.3 &81.3 &85.4 &87.3 &90.0 &96.6 &3.2e-02 \\
MCH &31.1$\pm$0.9 &29.6 &30.1 &31.1 &31.8 &32.7 &30.5$\pm$1.4 &27.0 &30.0 &30.7 &31.5 &33.1 &3.7e-02 \\
MCHC &34.6$\pm$0.9 &33.3 &34.1 &34.4 &34.7 &37.7 &34.9$\pm$0.9 &33.0 &34.2 &34.9 &35.4 &37.4 &3.4e-04 \\
RBC &4.8$\pm$0.4 &4.3 &4.5 &4.9 &5.2 &5.6 &4.8$\pm$0.3 &3.7 &4.5 &4.8 &5.0 &5.4 &3.7e-02 \\
RDW-SD &44.3$\pm$2.5 &39.1 &42.5 &44.5 &45.9 &48.8 &42.5$\pm$3.7 &36.6 &40.5 &41.5 &42.7 &55.3 &1.4e-07 \\
RDW-CV &13.4$\pm$0.6 &11.8 &13.0 &13.4 &13.9 &14.7 &13.2$\pm$1.2 &11.8 &12.5 &12.9 &13.5 &17.8 &1.5e-04 \\
WBC &7.1$\pm$2.3 &4.0 &5.2 &6.7 &8.5 &14.5 &6.4$\pm$1.9 &3.7 &5.0 &6.2 &7.5 &12.8 &3.6e-01 \\
IRF &10.3$\pm$3.2 &4.3 &8.4 &10.3 &11.6 &22.8 &9.3$\pm$3.4 &0.0 &6.9 &9.2 &11.5 &21.4 &1.3e-01 \\
LFR &88.9$\pm$3.5 &77.2 &87.2 &89.8 &91.1 &94.4 &91.0$\pm$3.0 &82.9 &88.8 &90.9 &93.3 &97.8 &1.5e-02 \\
MFR &9.4$\pm$2.3 &4.9 &8.0 &9.2 &10.5 &15.3 &7.9$\pm$2.5 &2.2 &5.9 &8.0 &9.6 &15.5 &7.3e-03 \\
HFR &1.7$\pm$1.5 &0.1 &0.9 &1.3 &1.8 &7.5 &1.1$\pm$0.6 &0.0 &0.6 &1.0 &1.6 &2.8 &2.9e-01 \\
OFF-HR &72.6$\pm$18.6 &31.9 &61.2 &73.8 &85.2 &107.8 &77.1$\pm$15.5 &25.7 &68.7 &76.1 &85.6 &111.7 &2.9e-02 \\
\cline{1-14}
\end{tabular*}
\label{table2}
\end{table*}

\section{Machine Learning Studies}\label{subsec3.3}

In this section, we described three machine learning algorithms that were trained and evaluated using the performance metrics. We used SVC, RF (tree-based ensemble) and XGBoost (tree-based boosting) algorithms in this analysis.

\subsection{ML algorithms}\label{subsec4.2}

\subsubsection{SVC}
Support vector machines (SVM) represent a class of supervised learning methods used for classification, regression and outlier detection \cite{bib13}. The main advantage of using SVM is its effectiveness in high dimensional spaces. In SVM, we construct a set of hyperplanes in a high dimensional space, that can be used to achieve a good separation between different classes of the dataset. An extension of SVM is kernel-based support vector classification (SVC), which can perform binary and multi-class classification on a dataset. Computing the outputs of SVC requires a kernel function, which takes two normalized input vectors as parameters and outputs a real number which characterizes how similar or dissimilar inputs are \cite{bib13}.

\subsubsection{RF}
Ensemble methods use multiple learning algorithms to obtain better predictive performance that could be obtained from any of the constituent learning algorithms alone. In other terms, it is a process of combining a finite number of weak learners in a more flexible structure to improve the prediction \cite{bib14}. Bagging is one of the ensemble methods where the prediction of each weak learner is assigned with equal weights, and the final classification is based on the voting of these predictions \cite{bib15}. Each weak learner is trained with a randomly drawn subset of the training samples where replacement is allowed. 

Random forest (RF) is a bagging method that combines several decision trees to improve the prediction \cite{bib16}. The main advantage of using RF is that the decision formed by taking several uncorrelated decision trees together outperforms the decision made by each individual decision tree. Furthermore, the low correlation between these weak learners is significant because it overcomes their individual errors \cite{bib16}.

\subsubsection{XGBoost}
Boosting is another example of ensemble methods, which involves building an ensemble by training each new learner on the data which is misclassified by the previous learner \cite{bib17}. In some cases, boosting has shown better performance than bagging, but it also tends to be more likely to overfit the training data. They are similar to a neural network with a single hidden layer, where the output is also a weighted sum of outputs of neurons \cite{bib17}. But they are trained differently because neural networks constitute adjustment of parameters of all neurons at once, whereas boosted algorithm consists of simple learners which are added sequentially one by one \cite{bib17}.

In this analysis, we used eXtreme Gradient Boosting (XGBoost) \cite{bib18}. It is a specific implementation of the gradient boosting algorithm, which uses second-order gradients of the loss function, which provides more information about the direction of gradients to get the minimum of the loss function. While regular gradient boosting uses the loss function of the base decision tree as a proxy for minimizing the error of the overall model, XGBoost uses the 2nd order derivative as an approximation. Another advantage of XGBoost is advanced regularization, which helps to overcome overfitting \cite{bib18}.

\subsection{Training}\label{subsec4.2}

We randomly partitioned the data samples such that 80\% of the data is used for training and 20\% for testing the algorithm. The data contains the 8 best haematological parameters, which are selected after performing the K-S test. The normalisation of the data is performed before training the algorithm.

SVC and RF algorithms are implemented using scikit-learn package \cite{bib19}, and XGBoost algorithm is implemented using xgboost package \cite{bib18}. The algorithms are trained using 5-fold cross-validation method.

Optimisation of these algorithms is performed using Optuna package \cite{bib20}. It allows us to define the hyperparameter space dynamically and efficient implementation of both searching and pruning strategies to find the best hyperparameter for the algorithm. 

\subsection{Evaluation metrics}\label{subsec4.2}

For binary classification, the evaluation of the algorithm during training can be defined based on the confusion matrix. In the confusion matrix, $TP$ and $TN$ denote the number of samples classified correctly by the algorithm as positive and negative, respectively, while $FN$ and $FP$ denote the number of misclassified positive and negative samples, respectively.

Accuracy is the most used evaluation metric, which tells us about the proportion of correct predictions made by the algorithm \cite{bib21}.
\[ Accuracy = \frac{TP+TN}{TP+FP+FN+TN} \]

But since we have a highly imbalanced dataset, it is also important to assess both the classes i.e. the amount of predicted positives that are actually positive (precision) and the amount of actual positives correctly classified (Recall). But since, there is a trade-off between precision and recall, so the best metric which maintain the balance between the two would be F1-score. It is a harmonic mean of both precision and recall. So, if either of the two is low, F1-score will be low and vice-versa \cite{bib21}.
\[ \text{F1-score} = \frac{Precision.Recall}{Precision + Recall} \]

Another metric we analysed is the Receiver Operating Characteristic (ROC) curve which describes the change in sensitivity with respect to (1-specificity) at different threshold values. The sensitivity tells us the proportion of correctly identified positives, and specificity measures the proportion of correctly identified negatives. 
\[ Sensitivity = \frac{TP}{TP + FN} \]
\[ Specificity = \frac{TN}{TN + FP} \]

The area under the ROC curve indicates how well the probabilities from the positive classes are separated from the negative classes. Therefore, it is also used as a metric for the evaluation of the algorithm \cite{bib21}.

\section{Results}\label{sec4}

\begin{figure}[hbt!]
\centering{\includegraphics[width=7cm]{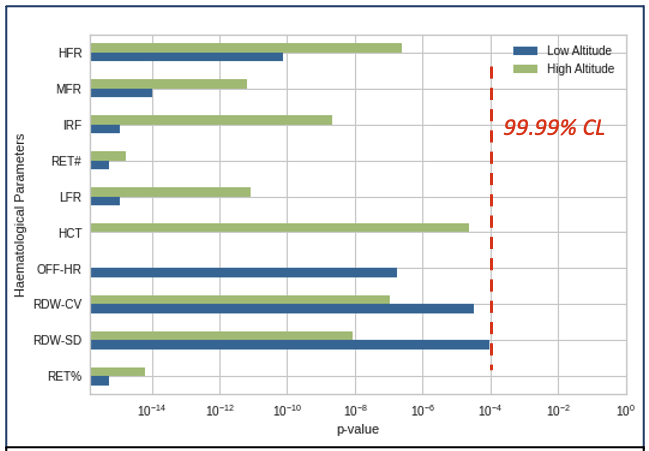}}
\caption{Best haematological parameters (at 99.99\% CL) from the K-S test at sea-level and high altitude.\label{result}}
\end{figure}

In this section, we reported the evaluation of both statistical and machine learning analysis. Fig. \ref{result} shows the results from the K-S test. The best haematological parameters for both sea-level and high altitude are shown with their distinguishing power independent of the model. It is observed that HCT is a good parameter to analyse the effect of rhEPO only at high altitudes, whereas OFF-HR shows the distinguishing power only at sea-level (low altitude).

In the next step, SVC, RF and XGBoost algorithms are trained, and the performance metrics are evaluated, as shown in Table \ref{tab3}. These algorithms use different approaches to predict the probability of a blood sample to be either a controlled sample or contained rhEPO. XGBoost achieves an accuracy of 92\%, which is higher than SVC (71\%) and RF(85\%). We used SOTA as the baseline to compare our results. In terms of sensitivity, XGBoost outperforms the SOTA with 68\%, whereas underperforms by 3\% in specificity. In general, both RF and XGBoost showed better performance than SVC. This is because both ensemble and boosting algorithms consist of more than one classifier for decision making. 

\begin{table}[htbp]
\caption{Comparison of evaluation result of three machine learning algorithms with SOTA}
\begin{center}
\begin{tabular*}{21pc}{@{\extracolsep{\fill}} c c c c c }
\hline
\textbf{Topic} & \textbf{SOTA} & \textbf{SVC} & \textbf{RF} & \textbf{XGBoost} \\
\hline
Accuracy &- &0.74 &0.85 &0.92 \\
F1-score &- &0.83 &0.92 &0.95 \\
Sensitivity &0.45 &0.46 &0.33 &0.68 \\
Specificity &1.0 &0.86 &0.98 &0.97 \\
AUC &0.84 &0.75 &0.84 &0.90 \\
\hline
\end{tabular*}
\label{tab3}
\end{center}
\end{table}

Another factor that could improve the result is by adding some domain knowledge of the haematological parameters in addition to the statistical results. Currently, the K-S test is used to select the best indicator, which is biased towards the data distribution. The presence of outliers in the data could possibly impact the $p$-values. Moreover, data statistics also play an important role, so the result can still be improved by increasing the number of samples to train the algorithm. 

\section{Conclusion}\label{sec5}

The direct laboratory-based methods are expensive, time-consuming process that needs the continuous monitoring of haematological parameters of each individual and require domain experts to analyse the blood samples. Therefore, there is a need for indirect methods of detection.   

In this paper, we present an indirect method to detect the presence of rhEPO in blood. We conducted a clinical experiment where the blood samples of 39 individuals (given placebo or rhEPO) were collected. We combined both statistical methods and machine learning algorithms to analyse the blood samples. The K-S test showed that the effect of rhEPO has more significance on RET\#. We set a threshold of 99.99\% confidence level where the 8 best indicators are selected for training the machine learning algorithms. We discarded the parameters which show high correlations. 

We trained SVC, RF and XGBoost algorithms on the blood samples and evaluated the algorithms by calculating the performance metrics. XGBoost algorithm showed better results and outperformed the SOTA method in all the metrics (except specificity). Our results suggest that the ensemble and boosting algorithms are effective for mapping the effect of rhEPO in haematological parameters. However, the result is limited to the amount of data available for performing this study. Improving data availability and data quality are potential keys to further enhance the performance of the algorithm. It also opens up the use of more sophisticated non-linear models like a neural network which often learns better with more statistics. There are data augmentation techniques like generative models that could possibly help to increase the data statistics.  

In general, AI-based algorithms have the potential to improve the current indirect methods in sports by using the insights from the data for better decision making. However, it is restricted by both the availability and confidentiality of the athletes' data. In this paper, we showed how the application of Artificial Intelligence offers a promising result and can contribute in a significant manner in improving the decision-making for the detection of drug-abused athletes in sports. In future, we aim to use generative models to increase the data statistics and train deep learning algorithms to improve the results.

\end{document}